\documentclass[10pt,twocolumn,letterpaper]{article}
\pdfoutput=1
\usepackage{iccv}
\usepackage{times}
\usepackage{epsfig}
\usepackage{graphicx}
\usepackage{amsmath}
\usepackage{amssymb}

\usepackage{mathtools}
\usepackage{amssymb}

\usepackage{array}
\usepackage{booktabs}
\usepackage{multirow}
\usepackage{subcaption}
\usepackage{float}
\usepackage{bbm}

\usepackage{stfloats}
\usepackage[breaklinks=true,bookmarks=false]{hyperref}

\iccvfinalcopy

\ificcvfinal\fi

\begin{document}

\title{Improving Binary Neural Networks through Fully Utilizing Latent Weights}

\author{Weixiang Xu$^{1}$ \qquad Qiang Chen$^2$ \qquad Xiangyu He$^1$ \qquad Peisong Wang$^1$ \qquad Jian Cheng$^1$\\
 \textrm{$^1$NLPR, Institute of Automation, Chinese Academy of Sciences} \qquad \textrm{$^2$Baidu Inc.}\\
{\tt\small \{xuweixiang2018@ia.ac.cn\}} \quad
{\tt\small \{chenqiang13@baidu.com\}} \quad \\
{\tt\small \{iva.shuanholmes@gmail.com\}} \quad
{\tt\small \{peisong.wang,jcheng@nlpr.ia.ac.cn\}}
}

\maketitle
\ificcvfinal\thispagestyle{empty}\fi

\begin{abstract}
    Binary Neural Networks (BNNs) rely on a real-valued auxiliary variable W to help binary training. However, pioneering binary works only use W to accumulate gradient updates during backward propagation, which can not fully exploit its power and may hinder novel advances in BNNs. In this work, we explore the role of W in training besides acting as a latent variable. Notably, we propose to add W into the computation graph, making it perform as a real-valued feature extractor to aid the binary training. We make different attempts on how to utilize the real-valued weights and propose a specialized supervision. Visualization experiments qualitatively verify the effectiveness of our approach in making it easier to distinguish between different categories.
    Quantitative experiments show that our approach outperforms current state-of-the-arts, further closing the performance gap between floating-point networks and BNNs. Evaluation on ImageNet with ResNet-18 (Top-1 63.4$\%$), ResNet-34 (Top-1 67.0$\%$) achieves new state-of-the-art.
\end{abstract}

\section{Introduction}

Binary neural networks (BNNs)~\cite{NIPS2016_d8330f85} have become one of the most popular topics for the deployment of computation-intensive deep convolutional neural networks on low-power devices. By constraining both weights and activations to $\pm 1$, BNNs benefit from up to $32\times$ compression ratio~\cite{NIPS2015_3e15cc11} and the replacement of expensive floating-point matrix multiplication by cheap bitwise xnor and popcount operations~\cite{NIPS2016_d8330f85}. However, the limitation of model capacity and approximate gradient make training accurate binary models more difficult than their full-precision counterparts~\cite{alizadeh2018empirical}, especially on large-scale datasets.

To solve this challenging problem, lots of constructive works focus on it from different perspectives, such as minimizing quantization error, designing architectures that are more suitable for binarization
, introducing additional regularization functions on binary weights or activations
, retaining information from optimizations viewpoint 
 and so on.

Despite these different perspectives, they commonly adopt the same training framework, in which a real-valued auxiliary variable $W$ is used to accumulate gradient updates during backward propagation and then binarized to $B$ at inference.
After training, $W$ is discarded, and $B$ will be deployed to the resource-limited device.

\begin{figure}[t]
\begin{center}
   \includegraphics[width=\linewidth]{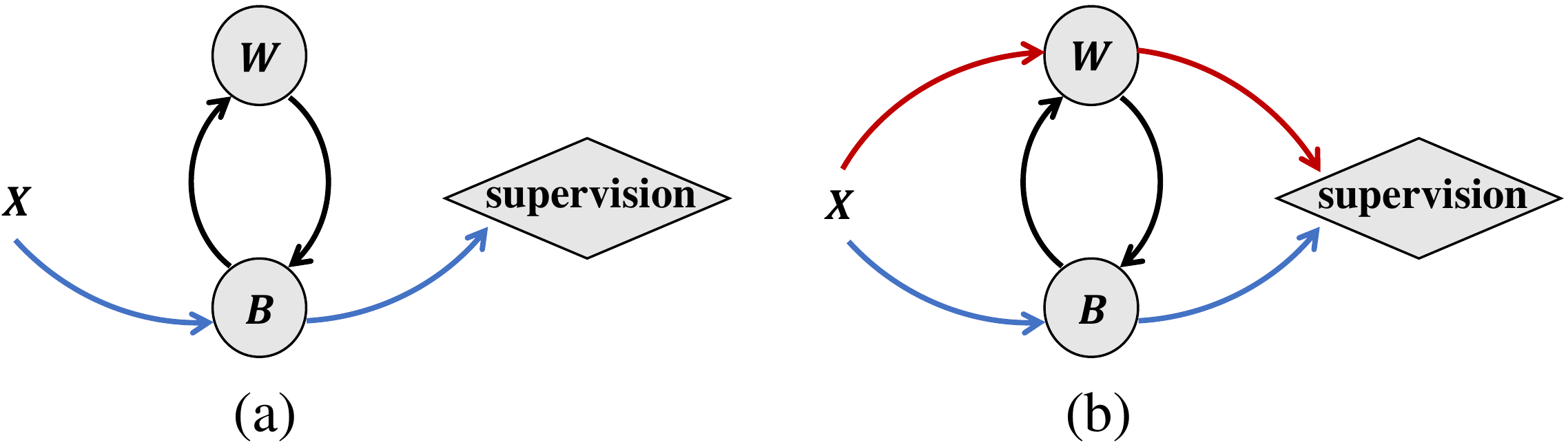}
\end{center}
   \caption{Comparison between pioneer binary training process with ours. (a) In previous works, the auxiliary variable $W$ serves as a latent variable to update the binary variable. (b) In this work, we propose to add $W$ into the computation paragraph, working as a real-valued feature extractor.}
\label{fig:introduction}
\end{figure}

\begin{figure*}[t]
\begin{center}
   \includegraphics[width=0.8\textwidth]{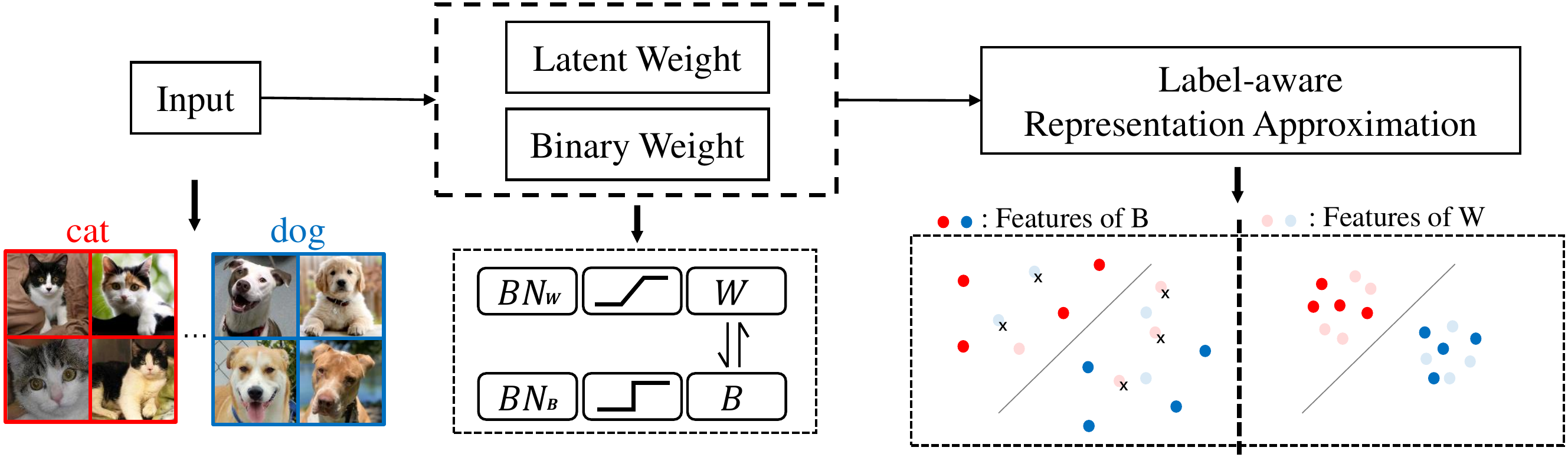}
\end{center}
   \caption{Overview of adding the latent weight into computation graph and being supervised by Label-aware
Representation Approximation. We use $BN_{\rm W}$ and $hard\_tanh(\cdot)$ to restore $W$'s ability as a real-valued feature extractor. `$\times$' in the figure indicates the samples misclassified by $W$. The specific supervision forces features of $W$/$B$ to present a similar distribution pattern and cluster more tightly with the same label.}
\label{fig:pipeline}
\end{figure*}

Different from previous works, where $W$ only acts as a latent variable that serves to update binary variable $B$, we aim to further explore the power of $W$. To begin with, we evaluate the network with floating-precision $W$ after training. A counterintuitive observation is that accuracy of $W$ on validation is extremely low, which contradicts our perception of it as a real-valued feature extractor. Nevertheless, we show that $W$ can restore its performance by simply renewing BN's running statistics and replacing $sign(\cdot)$ with real-valued activation $hard\_tanh(\cdot)$,
indicating that $W$ can extract helpful real-valued information. Based on this observation, we propose to add $W$ into the computation graph by letting $W$ participate in convolution with inputs such that features from $W$ can also be used for specialized supervision functions.
The penultimate features extracted by the network with $W$ and $B$ both contain high-level semantic information which is category-related. Therefore, in addition to the previous weight approximation, we propose our label-aware representation approximation by enforcing representations of the same class extracted by $W$ and $B$ to be close at the semantic level so that it is easier for BNNs to distinguish between different categories. We visualize penultimate features to verify our method's effectiveness and show that BNNs' performance can be further improved by fully utilizing latent weights $W$ during training. The overview of our training pipeline are illustrated in Fig.~\ref{fig:pipeline}.
Overall, this paper makes the following contributions:
\begin{itemize}
\item We revisit the pioneering binary training process and point out that the real-valued auxiliary variable $W$ has not been fully utilized. We propose to add $W$ into the computation graph as a real-valued feature extractor.
\item We propose label-aware representation approximation regularization which allows representations of $W$/$B$ to present a similar distribution pattern. Additionally, representations with the same label cluster more tightly, making it easier to distinguish between different categories.

\item Both qualitative and quantitative experiments are conducted to explore the effectiveness. In particular, experiments on ImageNet with ResNet-18 (Top-1 63.4$\%$), ResNet-34 (Top-1 67.0$\%$) achieve new state-of-the-art.
\end{itemize}
\section{Related Work}
Various methods have been proposed to reduce the parameter size and accelerate the inference phase, such as compact model architecture design~\cite{howard2017mobilenets,zoph2016neural}, quantization~\cite{gupta2015deep,zhou2016dorefa,cai2017deep}, network pruning~\cite{han2015deep_compression,he2019filter}, knowledge distilling~\cite{hinton2015distilling}, etc. In this work, we focus on binarization, the extreme case of quantization, which quantizes full-precision weights and activations to $1$ bit.
In order to narrow the accuracy gap between binary and full precision networks on large-scale data sets such as ImageNet, various influential works have emerged. (1) Following the way of approximating real-valued weights,~\cite{rastegari2016xnor,bulat2019xnor-net-plus} introduces scaling factors to minimize the quantization error.~\cite{lin2020rotated} proposes rotated matrix to approximate $W$ from both magnitude and direction perspective. Multi-bit methods~\cite{NIPS2017_b1a59b31,zhu2019binary,liu2019circulant,zhuang2019structured} propose decomposing a single convolution layer into K binary convolution operations, compensating for the information loss of binary weights. (2) Besides weight approximation, designing architectures that are more suitable for binarization also attracts increasing attention. Bi-Real~\cite{liu2020bi} proposes double residual connections with full-precision downsampling layers and XNOR++~\cite{bulat2019xnor-net-plus} uses PReLU to replace ReLU.~\cite{liu2020reactnet} replaces depthwise convolutions in MobileNet with vanilla convolutions to construct a new binary architecture. BATS~\cite{adrian_eccv20} applies NAS to search for accurate binary architectures. (3) Various regularization functions have been developed to either adjust weight distribution~\cite{darabi2018bnn+,bai2018proxquant,gu2019projection,gu2019bayesian} or to control the range of activations~\cite{ding2019regularizing}. (4) Some works make efforts to improve the performance of BNNs from the perspective of optimization~\cite{alizadeh2018empirical,han2020training,qin2020forward}, which modify propagation and retain the information in binary networks. Contrary to the specific binary optimizer Bop~\cite{NEURIPS2019_9ca8c9b0} which throws latent weight away, we aim to explore and exploit latent weights' untapped ability to improve the performance of BNNs in this work.

Almost all those BNNs works follow the same training pipeline. That is, using the sign function in forward propagation to binarize the floating-point auxiliary variable $W$ into $B$, and using Straight Through Estimator~\cite{bengio2013estimating} in backward propagation to pass gradient from $B$ to $W$. In previous binary training methods, real-valued weight $W$ only acts as a latent variable that serves to update binary variable $B$.
Unlike the above works, we focus on a new perspective that has been neglected by further exploring the role of $W$ in training and improving BNNs' performance through fully utilizing $W$.
\section{Approach}
In this section, we first briefly introduce the formulation of training BNNs and our motivation to utilize the latent weights to help improving BNNs. Then we introduce a simple but effective approach on how to utilize latent weights. Besides that, we further explore our methods' effectiveness, including aspects of representation visualization and feature reconstruction.

\subsection{Background and Motivation}
\label{sec:motivation}
The gradient-based method of training BNNs, proposed by~\cite{NIPS2015_3e15cc11,rastegari2016xnor} at first, uses Straight Through Estimator (STE) to tackle the non-differentiable problem in binarization training, and has been adopted extensively by almost all subsequent works. Regardless of their differences, a real-valued auxiliary variable $W$, also known as latent weight, is commonly used to assist in training binary variable $B$ in the STE-based framework. Concretely, during the forward pass, $B$ is obtained by binarizing $W$:
\begin{equation}
\label{WandB}
        B = sign(W)
\end{equation}
along with a scale factor $\alpha$ which is obtained by minimizing quantization error $\Vert W- \alpha B \Vert^2_2$. Then they are used to perform $\alpha B \circledast sign(X)$, where $\circledast$ denotes the binary convolution and X is the input feature map. Moreover, during the backward pass, the gradient of $W$ will be computed by STE with approximation as:
\begin{equation}
\label{STE}
        \frac{\partial \mathcal{L}}{\partial W} \approx \frac{\partial \mathcal{L}}{\partial B} \cdot \mathbbm{1}_{|W|<1}
\end{equation}
where $\mathcal{L}$ is the task-related loss function, such as cross-entropy loss.

Through analyzing the whole training process, we summarize two roles of latent weight $W$ in the traditional binary training pipeline as below: (1) During the forward propagation, $W$ provides binary variable $B$ and floating-point scaling factor $\alpha$ by solving the approximation problem of minimizing quantization error $\Vert W- \alpha B \Vert^2_2$. (2) During the backward propagation, $W$ updates its value with approximated gradient obtained from STE and prepares for the next iteration of forward pass.

Since $B$ is the binary counterpart of the real-valued $W$, it is intuitive that the accuracy of a model with $W$ should be higher than that with $B$. However, the actual evaluation shows that this is not the case. We train a compact ResNet-18 using Bi-Real~\cite{liu2020bi} on CIFAR-10 and evaluate the accuracy of $W$ and $B$ separately. As can be observed from Table.~\ref{tab:W_B_accuracy}, the accuracy evaluated with $W$ is much lower than that of $B$ ($41.13\%$ vs. $82.37\%$).
We attribute this phenomenon to the bias in the statistics memorized by batch normalization (BN) in BNNs:
To decrease binarization's information loss, a BN layer is commonly inserted before the zero-threshold binarization function $sign(\cdot)$~\cite{rastegari2016xnor}, normalizing the real-valued features to zero mean. During training, the statistics (running mean $\mu$ and running variance $\sigma^2$) memorized by BN are calculated as:
\begin{equation}
\label{mu}
        \mu = \frac{1}{N} \sum_{i=1}^{N} F^i,\quad \sigma^2 = \frac{1}{N} \sum_{i=1}^{N} (F^i - \mu)^2
\end{equation}
where $F^i=\alpha B \circledast sign(X^i)$ is extracted by the binary weights.
So if we utilize $W$ instead of $B$ to perform convolution but still use the statistics inherited from outdated BN, the accuracy will degenerate.

\begin{figure}[t]
\begin{center}
   \includegraphics[width=0.95\linewidth]{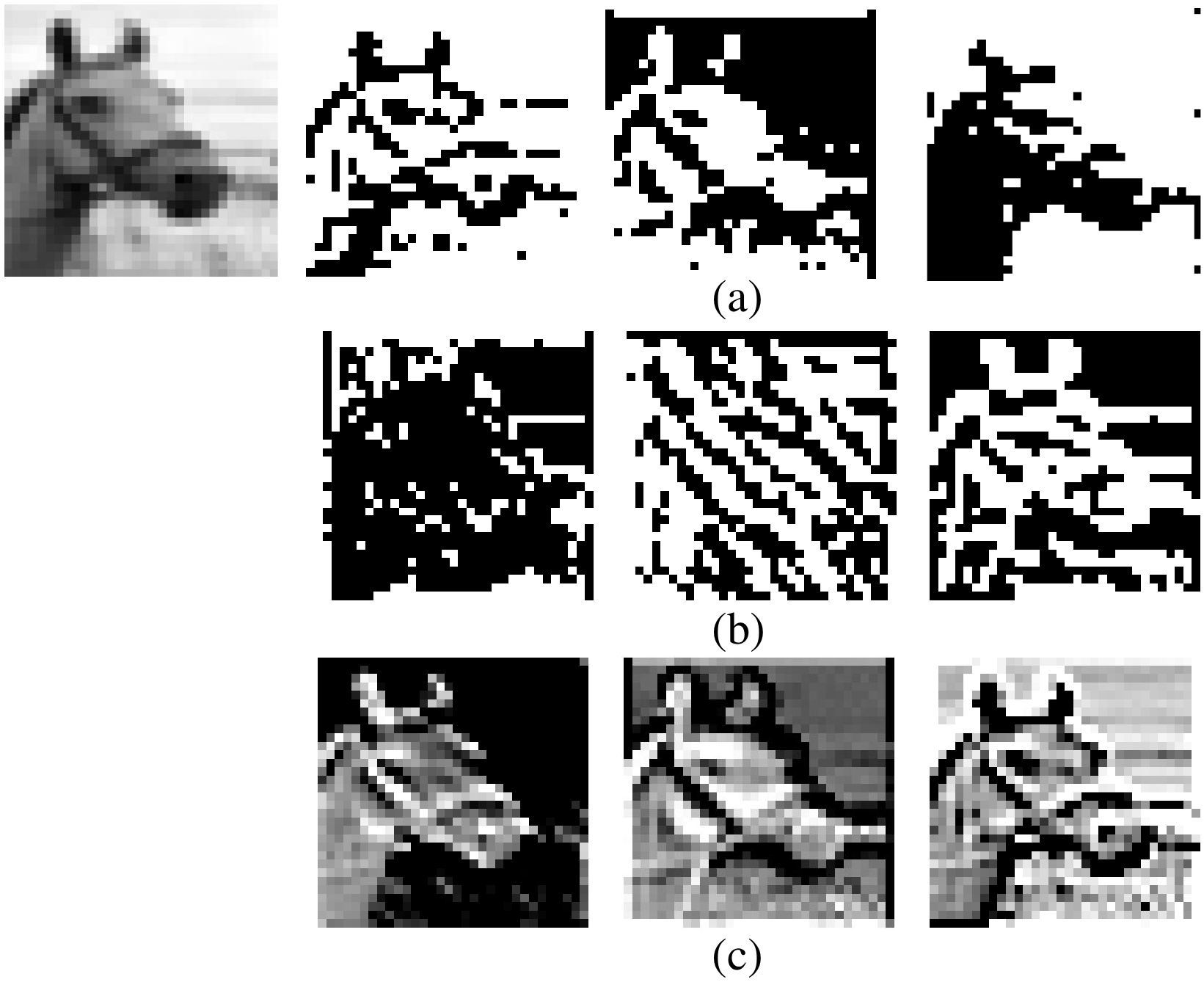}
\end{center}
   \caption{Visualization of intermediate feature maps after binary activation. (a) The binary features obtained with $B$ and its own BN. (b) The binary features obtained with $W$ and outdated BN
   . (c) Features of the same layer obtained with $W$ after recalculating BN and replacing with $hard\_tanh(\cdot)$.}
\label{fig:compare_three_w}
\end{figure}

\begin{table}[t]
\begin{center}
\begin{tabular}{c|cccc}
\hline
ResNet-18     &$B$   &$W$      &$W^{\dagger}$      \\
\hline
Top-1($\%$)   &82.37 &41.13    &80.89  \\
\hline
\end{tabular}
\end{center}
\caption{Accuracy of compact ResNet-18 (kernel stage 16-16-32-64) using Bi-Real~\cite{liu2020bi}. Evaluation with $B$/$W$. Here $\dagger$ indicates recalculating BN and replacing $sign(\cdot)$ with $hard\_tanh(\cdot)$.}
\label{tab:W_B_accuracy}
\end{table}

Based on the above analysis, this accuracy degeneration phenomenon can be mitigated by simply recalculating the BN layer statistics.
Apart from renewing BN, motivated by the fact that gradient of $sign(\cdot)$ at back-propagation is approximated by $hard\_tanh(\cdot)$ as Eq.\ref{STE}, we propose to replace binary activation with $hard\_tanh(\cdot)$ when preforming inference with $W$, recovering some feature information.
To illustrate it, we visualize the intermediate feature after binarization as Fig.~\ref{fig:compare_three_w}. The first raw shows the binary features obtained with $B$. Features in the second raw are obtained with $W$ and outdated BN, which have significant information bias.
The third raw are features of the same layer obtained with $W$ after renewing BN's statistics and replacing $sign(\cdot)$ with $hard\_tanh(\cdot)$. There are significant improvements compared with the second raw and different texture information compared with the first raw. The corresponding accuracy is shown in Table~\ref{tab:W_B_accuracy}. By means of the above processing, $W$ can restore the evaluation accuracy to be comparable with $B$, which is consistent with the above analysis. Although there is still a little gap between the evaluation with $B$ and $W$, the comparable performance indicates that features with different details shown in Fig.~\ref{fig:compare_three_w}c contain additional useful information.

From the above analysis, we know that the latent weight $W$ does not directly perform convolution with feature maps in the traditional binary training framework, neglecting its capability as a real-valued feature extractor. To this end, we make a step further towards this new perspective by investigating the following question: how to utilize the real-valued $W$ to improve binary training?
\subsection{Beyond Weight Approximation}
\label{sec:detail_approach}
In this section, we provide a feasible solution to the above question. We propose to add $W$ to the computation graph, which can introduce more details than the binary features extracted by $B$. Then we propose a simple but effective method to fully utilize the information extracted by $W$.

\subsubsection{Inference with Latent Weights}
\label{sec:Inference}
According to the analysis in section~\ref{sec:motivation}, recalculating BN's statistics and replacing $sign(\cdot)$ with real-valued activation $hard\_tanh(\cdot)$ have a significant impact on the performance of $W$. During training, instead of renewing BN's statistics after training, we use two sets of $\mu, \sigma^2$ for $W$ and $B$ to record the layer-wise statistics as Eq.~\ref{mu}.
We use $Y$ and $\widetilde{Y}$ to represent the features that obtained by $B$ and $W$ respectively.
For a network with $L$ layers, the $l$-th layer's feature $Y_l, \widetilde{Y}_l$ are calculated using the same `BinAct-Conv-BN-Activation' inference block as below:
\begin{equation}
    \begin{cases}
    Y_l = \Psi(BN_{\rm B} (\alpha B\circledast sign(Y_{l-1})))  \\
    \widetilde{Y}_l = \Psi(BN_{\rm W} (W\ast hard\_tanh(\widetilde{Y}_{l-1})))  \\
    \end{cases}
\label{eq:inference_w}
\end{equation}
where $B=sign(W)$ and $\Psi (\cdot)$ represents non-linear activation function. $BN_{\rm W}$ and $BN_{\rm B}$ share the same learnable affine coefficients but maintain their respective running statistic ($\mu_{W}, \sigma^{2}_{W}$ vs. $\mu_{B}, \sigma^{2}_{B}$) individually.

It is intuitive that if $W$'s accuracy is improved, its binary counterpart $B$'s accuracy will also be improved because $B$ is obtained from $W$ by weight approximation.
And the most straightforward idea is directly supervising $W$ with another cross-entropy loss $\ell_W$ on the last layer's output prediction $\widetilde{Y}_L$.
However, we find that it can not converge in practice. We attribute it to the fact that gradients $\partial \ell_W/\partial W$ derived from the new supervision dominate the update of real-valued weights compared with the original gradients from Eq.~\ref{STE}. However, weights optimized in the real-valued domain are not optimal in the binary domain, resulting in optimization oscillation. Therefore, we detach the gradients with respect to $\widetilde{Y}$ from the computation graph
in the following experiments.
\begin{figure*}[t]
\centering
\subcaptionbox{\label{subfig:a}}
    {%
        \includegraphics[width = .165\textwidth]{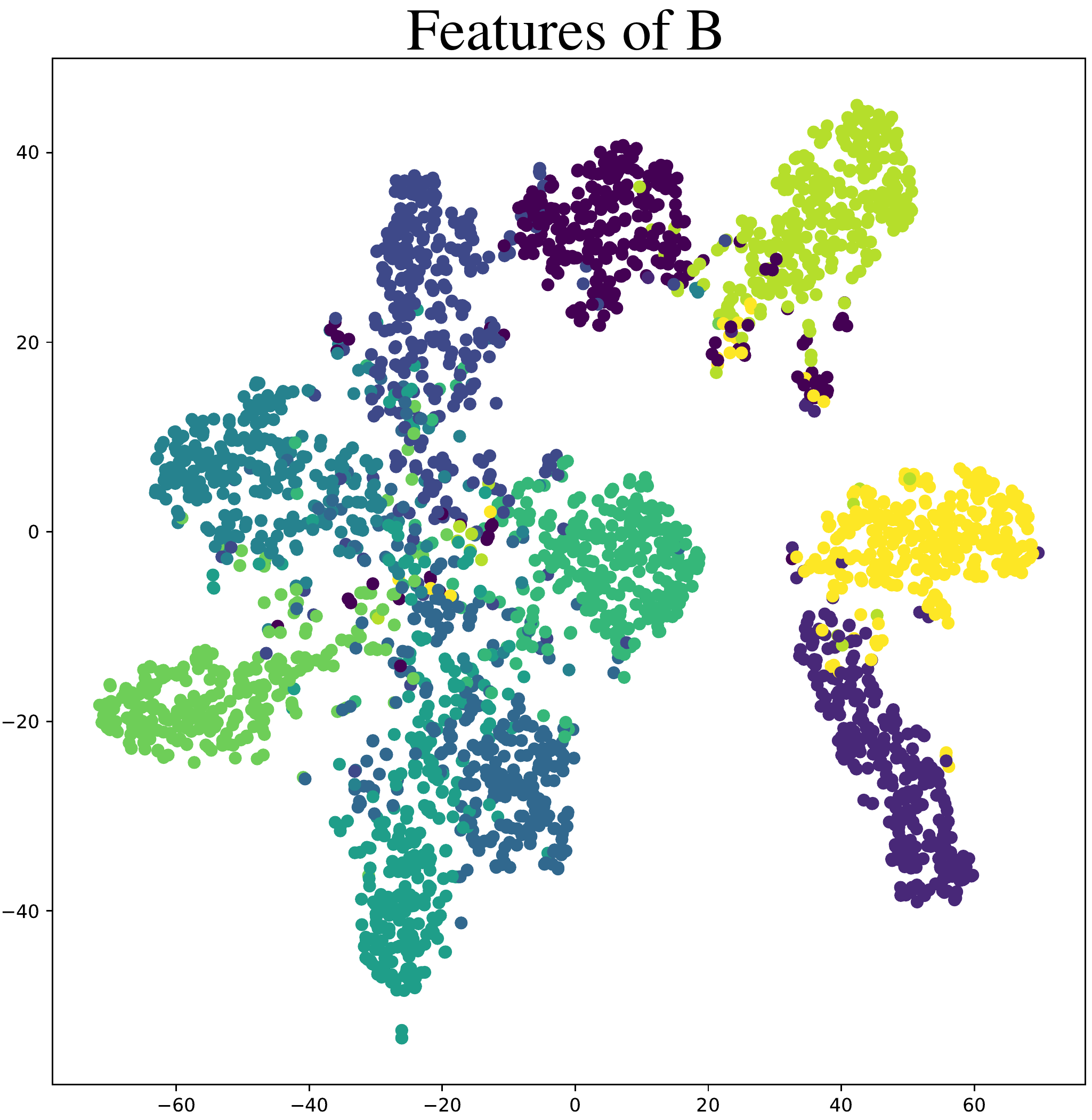}
        \includegraphics[width = .165\textwidth]{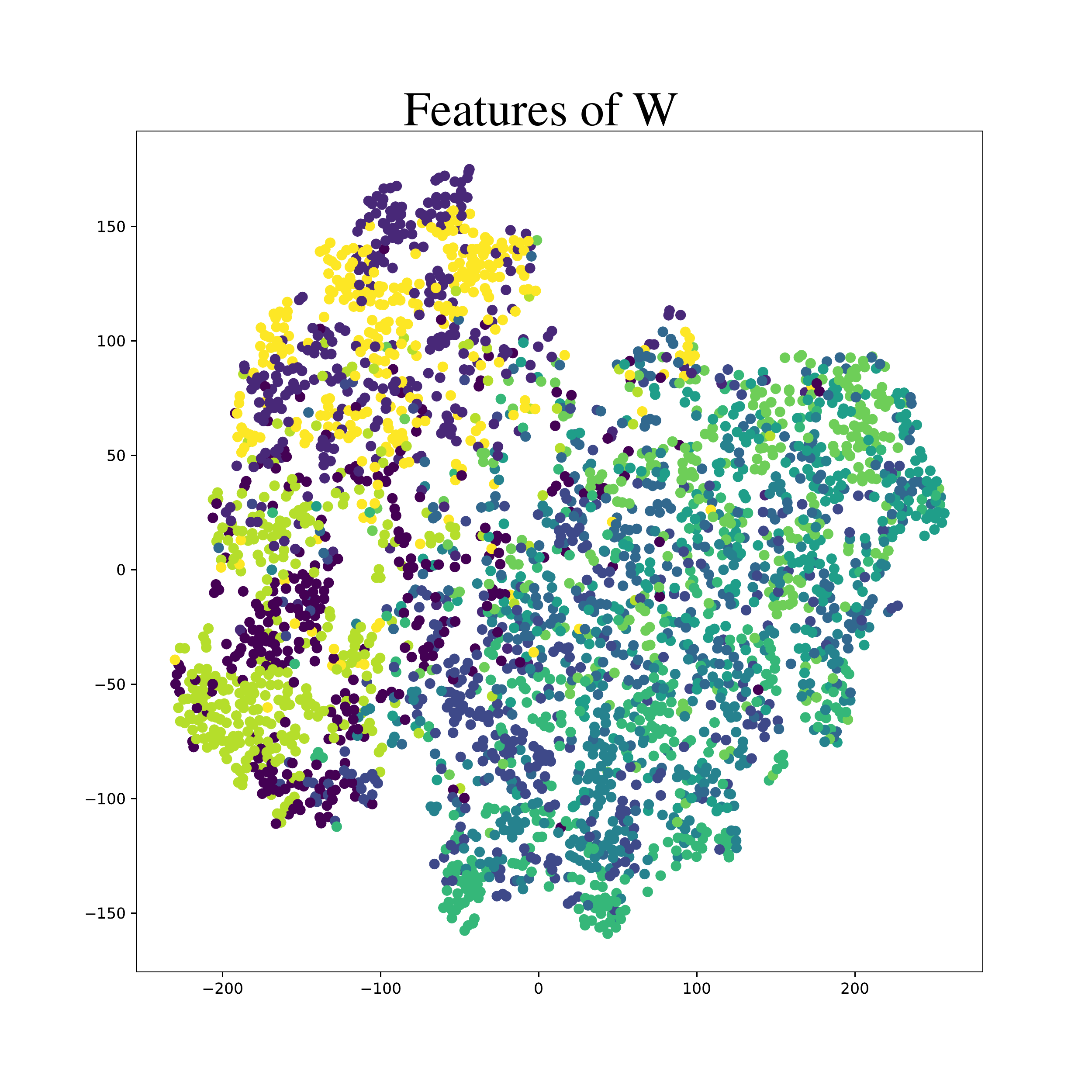}}
\subcaptionbox{\label{subfig:b}}
    {%
        \includegraphics[width = .165\textwidth]{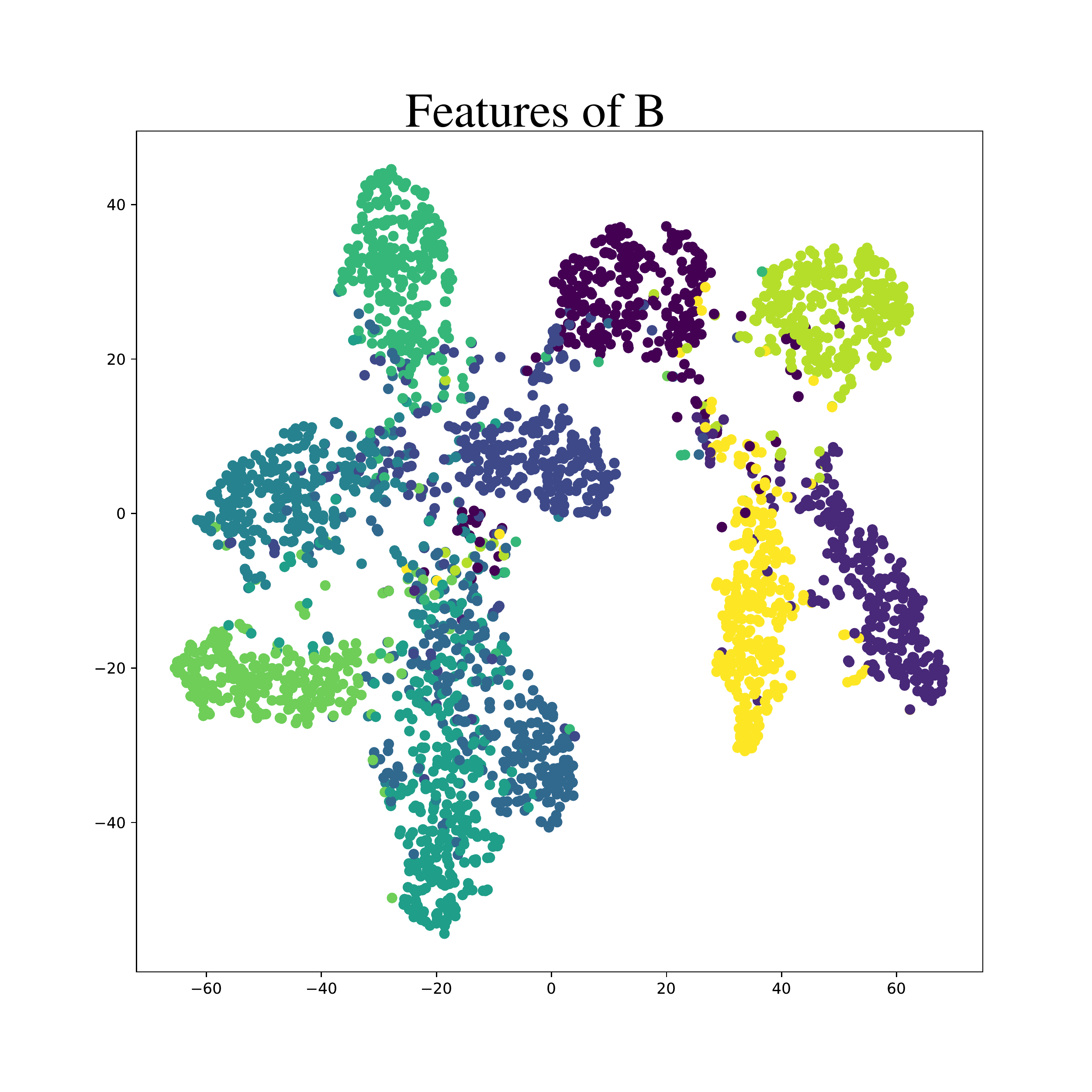}
        \includegraphics[width = .165\textwidth]{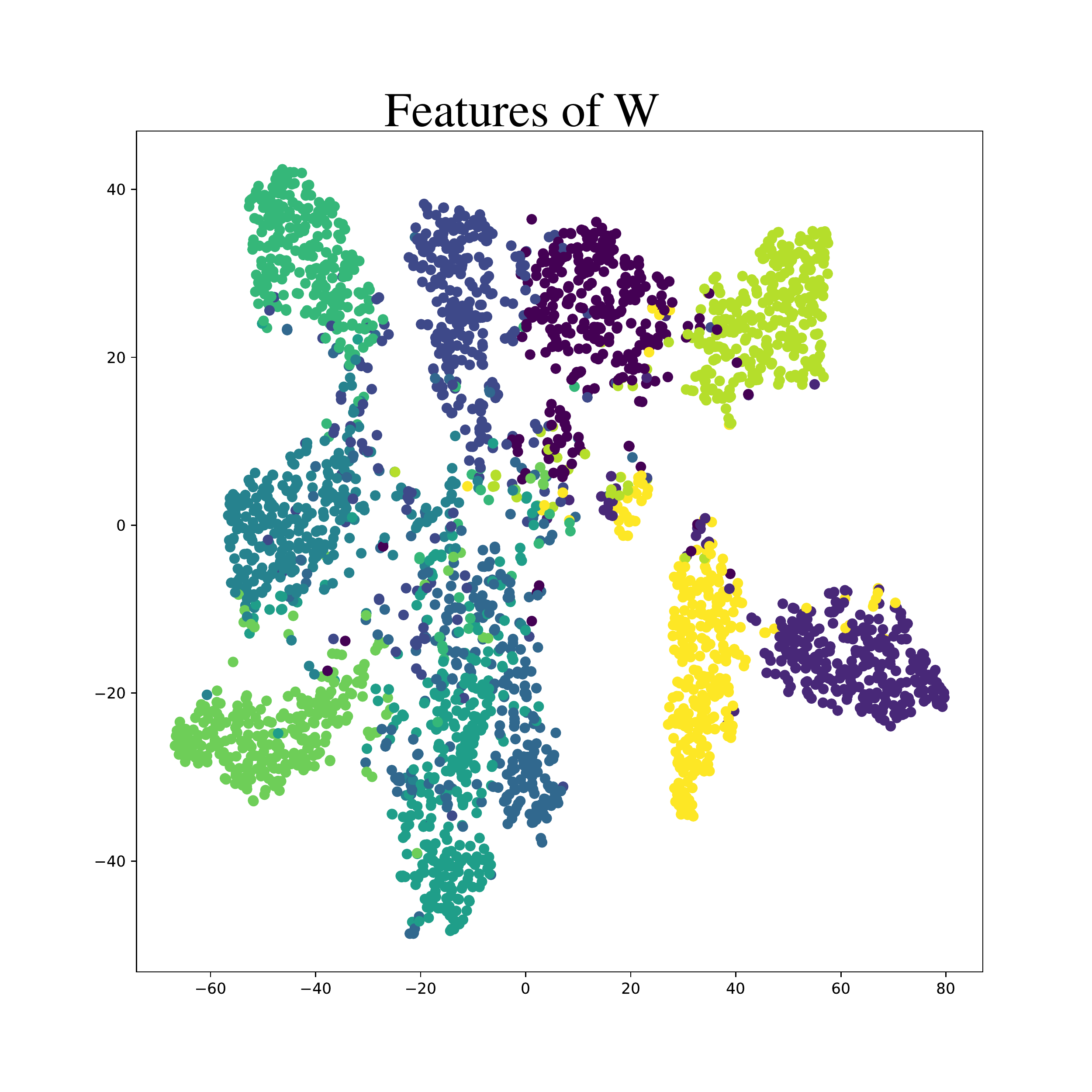}}
\subcaptionbox{\label{subfig:c}}
    {%
        \includegraphics[width = .165\textwidth]{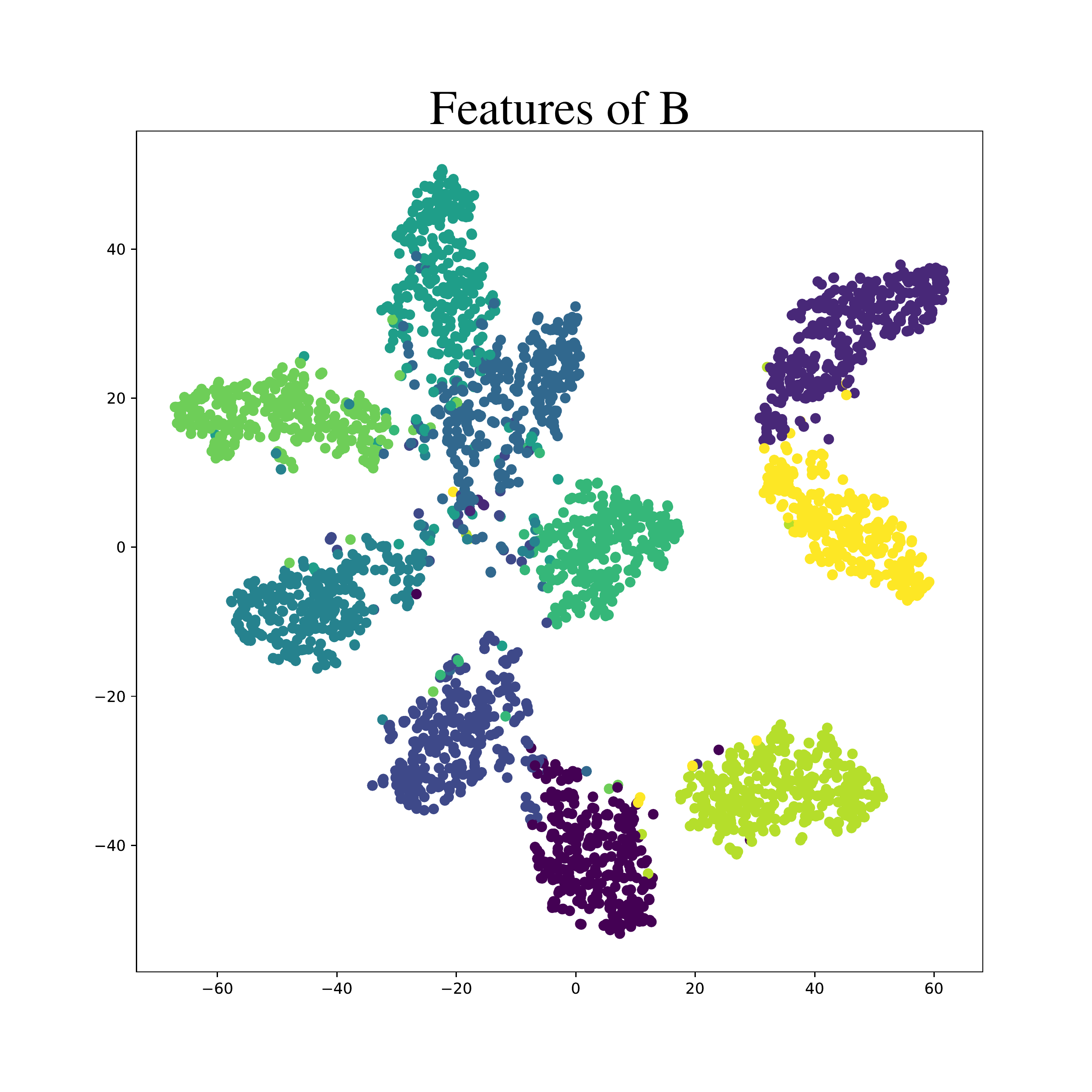}
        \includegraphics[width = .165\textwidth]{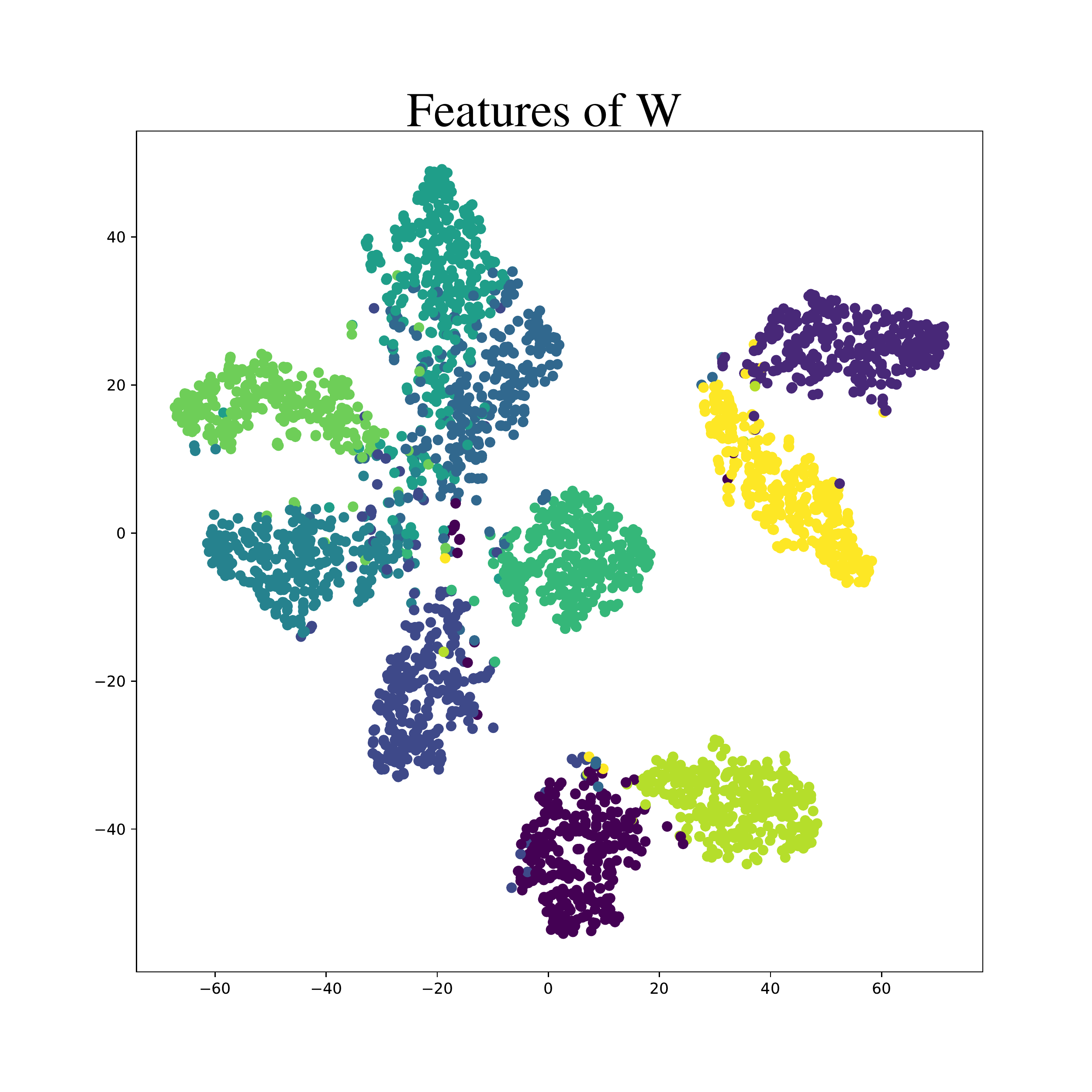}}
\caption{Penultimate layer representation visualization obtained by $W$/$B$ with three different training strategies. (a) We use Bi-Real as a benchmark. Here, latent weight $W$ does not perform inference and only serves to update the binary variable. (b) Training with Bi-Real + `instance-level' approximation Eq.~\ref{eq:representation_approximation}. The latent weight's feature extraction ability is restored. The aligned features present a similar distribution pattern.
(c) Training with Bi-Real + `category-level' approximation Eq.~\ref{eq:new_representation_approximation2}. Representations with the same label cluster more tightly on both $W$ and $B$.
\label{fig:visual_representation}}
\end{figure*}
\subsubsection{Label-aware Representation Approximation}
In the above section, we add the latent weight to the computation graph and obtain two different features $\widetilde{Y}$, $Y$.
Both features are extracted with the same network architecture using Eq.~\ref{eq:inference_w}. The difference is that one is conducted by the real-valued weight $W$ and real-valued activation $hard\_tanh(\cdot)$ while the other by the binary counterpart $B$ and $sign(\cdot)$.
As illustrated in Fig.~\ref{fig:compare_three_w}, the latent weights' features $\widetilde{Y}$ focus on different details, which can be combined with the original binary features $Y$ to assist binary training.

During forward propagation, different details in $\widetilde{Y}_l$ and $Y_l$ are conducted by $W$ and $B$ across multiple layers and will be aggregated in the penultimate layer features $\widetilde{Y}_{L-1}$ and $Y_{L-1}$. Then they will be fed into the same linear classifier for classification.
$\widetilde{Y}_{L-1}$/ $Y_{L-1}$ contain high-level semantic information extracted by the real-valued/ binary neural network, which are highly correlated with the classification targets. For the same input image, $\widetilde{Y}_{L-1}$ encodes different details with additional valuable information compared with $Y_{L-1}$. In this work, we use $\widetilde{Y}_{L-1}$ to provide extra supervision for improving BNNs' performance. For simplicity, we omit the subscript and just use $\widetilde{Y}$/ $Y$ to represent the penultimate layer features in the following.

To begin with, given a batch of images $\{X^i\}_{i=1...N}$, we first perform naive representation approximation, pushing $Y$ close to $\widetilde{Y}$ by minimizing the following formulation:
\begin{equation}
    \sum_i^N  \Phi_{\widetilde{Y}^i,Y^i} =
    \sum_i^N  \Vert \widetilde{Y}^i - Y^i \Vert^2_2
\label{eq:representation_approximation}
\end{equation}
Here, we use $\Phi_{p,q}$ to represent $\ell_2$ distance between two vectors $p$ and $q$.
In this way, representations $\widetilde{Y}$ and $Y$ from the same image will be forced to be closer. Namely, representation approximation is conducted image by image on the \textit{`instance level'}.
However, the instance-level approximation may not fully exploit the category information encoded by the representations.
In other words, instance-level approximation aligns the representations extracted by the binary backbone and the latent backbone for each image separately, which does not take advantage of the label information in the instance itself.

Considering the category information encoded by penultimate representations between different instances, we further introduce label supervision by pulling $\widetilde{Y}$ and $Y$ with the same label closer to each other on the basis of Eq.~\ref{eq:representation_approximation}. Given a batch of images with labels $\{X^i,T^i\}_{i=1...N}$, we formulate our label-aware representation approximation loss $\mathcal{L}_{rep}$ as below,

\begin{equation}
    \sum_i^N K \Big[\underbrace{\Phi_{\widetilde{Y}^i,Y^i} }_{\text{Instance level}} +
     \underbrace{\sum_{j \in I(i)}(
    \Phi_{Y^i,Y^j}
    +\Phi_{\widetilde{Y}^i,Y^j}
    + \Phi_{Y^i,\widetilde{Y}^j})}_{\text{Category level}} \Big]
\label{eq:new_representation_approximation2}
\end{equation}
where $I(i)$ is the index $k$ of $\widetilde{Y}^k$ or $Y^k$ in the current batch ($k=1...N$ and $k\neq i$) that has the same category with $\widetilde{Y}^i$. $|I(i)|$ is the number of elements to be indexed.
The averaging factor $K=\frac{1}{(3|I(i)|+1) |\widetilde{Y}^i|}$ is a constant to normalize the loss.

In Eq.~\ref{eq:new_representation_approximation2}, the item $\Phi_{\widetilde{Y}^i,Y^i}$ is inderited from Eq.~\ref{eq:representation_approximation} to align representations.
Besides,
$\Phi_{Y^i,Y^j}$ , $\Phi_{\widetilde{Y}^i,Y^j}$ and $\Phi_{Y^i,\widetilde{Y}^j}$ pulls the aligned representations together in pairs, which symmetrically reduces intra-class features variations.
Among of them, the item $\Phi_{Y^i,Y^j}$ only uses the binary features extracted by $B$ and reduces intra-class distance by making binary features of the same class close to each other, which provides similar capability to the well-known center loss~\cite{wen2016discriminative}. The other two items $\Phi_{\widetilde{Y}^i,Y^j}$ and $\Phi_{Y^i,\widetilde{Y}^j}$ further introduce category-related information encoded by $\widetilde{Y}$, which supervises the binary training with the latent features that focus on different details.
In this way, we achieve representation approximation from instance-level to \textit{category-level}. Note that
$\Phi_{\widetilde{Y}^i,\widetilde{Y}^j}$
is not included in the loss term because we detach the gradient from $\widetilde{Y}$ as analyzed in section~\ref{sec:Inference}.

The label-aware representation approximation loss $\mathcal{L}_{rep}$ in Eq.~\ref{eq:new_representation_approximation2} can be optimized through the gradient descent and the gradient with respect to $Y^i$ can be calculated as:
\begin{equation}
\resizebox{.9\linewidth}{!}{$
\begin{aligned}
    \frac{\partial \mathcal{L}_{rep}}{\partial Y^i} =
    &2K \Big[(Y^i-\widetilde{Y}^i)+
    \sum_{j \in I(i)} \Big(2Y^i- Y^j-\widetilde{Y}^i
    \Big)\Big] \\
    =&2K \Big[
    (2|I(i)|+1)Y^i - \sum_{\mathclap{j\in \{I(i)+i\}}}\widetilde{Y}^j - \sum_{\mathclap{j \in I(i)}} Y^j
    \Big]
\end{aligned}
$}
\label{eq:grad}
\end{equation}

Based on the gradient chain rule, we further have,
\begin{equation}
\begin{aligned}
    \frac{\partial \mathcal{L}_{rep}}{\partial W}= \sum_i^{N}
                \frac{\partial \mathcal{L}_{rep}}{\partial Y^i}
                \frac{\partial Y^i}{\partial B}
                \frac{\partial B}{\partial W}
\end{aligned}
\label{eq:grad_w}
\end{equation}
where $\frac{\partial Y^i}{\partial B}$ and $\frac{\partial B}{\partial W}$ can be calculated through Eq.~\ref{eq:inference_w} and~\ref{STE} respectively. The complete loss function $\mathcal{L}$ is a linear combination of the standard cross-entropy loss $\mathcal{L}_{ce}$ and representation approximation loss $\mathcal{L}_{rep}$:
\begin{equation}
\begin{aligned}
    \mathcal{L}=\mathcal{L}_{ce}+\lambda \mathcal{L}_{rep}
\end{aligned}
\label{eq:loss}
\end{equation}
$\lambda$ is a balance factor on $\mathcal{L}_{rep}$ which regularizes the extent of representation approximation as well as pulls the binary representation from the same class closer.

\subsection{Effectiveness Exploration}
In this section, we explore the effect of the proposed method from a qualitative view. Precisely, we first visualize the penultimate layer's representations extracted by the binary kernel $B$ and corresponding real-valued latent kernel $W$ respectively. Then we discuss the relationship between the feature approximation error and classification error.

\noindent\textbf{Representation Visualization}
The penultimate layer's representation contains rich semantic information extracted by the backbone as well as category-related information that will be fed to the linear classifier for classification.
We visualize the penultimate layer representations of ResNet-18 (kernel stage 16-16-32-64) trained on CIFAR-10 with t-SNE~\cite{van2008visualizing} in Fig.~\ref{fig:visual_representation}.
Among them, we compare three different binary training strategies: (1) The baseline trained with Bi-Real~\cite{liu2020bi} in which latent weight $W$ only serves to update the binary variable. (2) Based on (1), we further perform inference with latent weights, thus obtaining features $\widetilde{Y}$ extracted by $W$. Then regularization as Eq.~\ref{eq:representation_approximation} that forces $\widetilde{Y}$ and $Y$ from the same image closer is added. (3) Based on (2), we further take category information into consideration and use Eq.~\ref{eq:new_representation_approximation2} as our final regularization.

Representations of both $B$ and $W$ obtained from the three strategies above are shown in Fig.~\ref{fig:visual_representation}. Their quantitative results in terms of accuracy are reported in ablation study.
Comparing Fig.~\ref{subfig:a} and \ref{subfig:b}, although the added representation approximation constraint Eq.~\ref{eq:representation_approximation} makes the representations obtained by $W$ and $B$ look more similar on the 2D visual distribution map, it does not help to improve the classification accuracy.
Comparing Fig.~\ref{subfig:b} and \ref{subfig:c},
by pulling representations $\widetilde{Y}$, $Y$ of different images with the same labels together, representation of $B$ in Fig.~\ref{subfig:c} exhibits more compact clustering within the same categories and is more easily separated by the classifier.

\noindent\textbf{Feature Reconstruction vs. Classification Accuracy}
In Xnor-Net~\cite{rastegari2016xnor}, a real-valued factor $\beta$ is introduced for compensating the error between the features obtained by binary convolution and those obtained by full precision convolution. 
Here we define feature reconstruction error in the $l$-th layer as,
\begin{equation}
\begin{aligned}
    R_l(\widetilde{Y}_l, Y_l) = \frac{1}{N\cdot C\cdot H\cdot W}\Vert \widetilde{Y}_l - Y_l \Vert^2_2
\end{aligned}
\label{eq:reconstruct_loss}
\end{equation}
where, $\widetilde{Y}_l, Y_l \in \mathbb{R}^{N\times C\times H\times W}$ are features obtained by latent weight $W$ and binary weight $B$, respectively.
A direct way to reduce reconstruction errors is to add reconstruction error regularization $\sum_{l=1}^{L}R_l(\widetilde{Y}_l, Y_l)$ (Eq.~\ref{eq:reconstruct_loss}) in addition to the classification loss, where $L$ is the number of convolution layers. We still use the compact ResNet-18 with Bi-Real~\cite{liu2020bi} and PReLU~\cite{bulat2019xnor-net-plus} as our benchmark. Different penalty coefficients for the reconstruction error regularization have been tested, and we report the highest accuracy result. Table~\ref{reconstruction_error} details the relationship between layer-wise feature reconstruction error and the final accuracy. Note that simply minimizing local layer-wise feature reconstruction error with `MIN FRE' generates near 7$\times$ smaller average feature reconstruction error than the baseline, yet resulting in worse accuracy. We believe there are two reasons for the failure of directly minimizing the reconstruction error: First, the binary weight and its corresponding real-valued weight are mutually coupled during training and it is easy to fall into a local collapsing solution by directly minimizing reconstruction error. For example, $W = B=0$ is one of the extreme cases. Second, minimizing local reconstruction error does not consider classification-related categories. Minor reconstruction errors do not necessarily lead to better classification results. We show that our method presents a trade-off between feature reconstruction loss and the global cost function and achieves the highest performance.
\begin{table}[t!]
\caption{Layer-wise feature reconstruction error and final accuracy. We use compact ResNet-18 
on CIFAR-10 as the baseline. `MIN FRE' refers to minimizing feature reconstruction error.}
\label{reconstruction_error}
\begin{center}
\resizebox{0.9\columnwidth}{!}{
\begin{tabular}{c|ccc}
\toprule[1.pt]
\multirow{2}{*}{Layer} & \multirow{2}{*}{Baseline} & Baseline  & Baseline \\
& & +Min FRE & +Ours  \\
\toprule[1.pt]
Conv2 & 0.00541 & 0.00088 & 0.00186    \\
Conv3 & 0.00968 & 0.00121 & 0.00291    \\
Conv4 & 0.01322 & 0.00145 & 0.00361    \\
Conv5 & 0.01462 & 0.00157 & 0.00427    \\
Conv6 & 0.01673 & 0.00175 & 0.00529    \\
Conv7 & 0.01715 & 0.00194 & 0.00601    \\
Conv8 & 0.01991 & 0.00217 & 0.00691    \\
Conv9 & 0.02136 & 0.00240 & 0.00823    \\
Conv10 & 0.02531 & 0.00338 & 0.00866    \\
Conv11 & 0.03109 & 0.00390 & 0.00989    \\
Conv12 & 0.03478 & 0.00446 & 0.01167    \\
Conv13 & 0.04134 & 0.00489 & 0.01237    \\
Conv14 & 0.05082 & 0.00695 & 0.01762    \\
Conv15 & 0.06010 & 0.00899 & 0.02097    \\
Conv16 & 0.06300 & 0.01211 & 0.02525    \\
Conv17 & 0.09141 & 0.01938 & 0.03669  \\
\textbf{Average} &0.03225  & \textbf{0.00484}  &0.01138  \\
\hline
\textbf{Acc. (\%)} &82.37  & 69.25  & \textbf{85.10} \\
\toprule[1.pt]
\end{tabular}}
\vspace{-5mm}
\end{center}
\end{table}

\section{Experiments}
In this section, we evaluate the proposed method in terms of accuracy and efficiency. Our experiments are conducted on CIFAR-10/100 and ImageNet datasets. Several mainstream networks such as VGG, ResNet and Wide ResNet (WRN) are used for testing.

\subsection{Implementation Details}
\label{sec:baseline}
\noindent\textbf{Network Structure.} It has been well known that the optimal architecture for full precision networks and binary networks is different~\cite{rastegari2016xnor,alizadeh2018empirical,liu2020bi,martinez2020training,bulat2020high}. As almost all previous binary works, we keep the first and last layers real-valued in all experiments. For VGG, we use the same architecture with~\cite{NIPS2015_3e15cc11,rastegari2016xnor} in which a dropout layer with $p = 0.5$ is placed before the last layer. For ResNet and WRN, the modified architectures used in this work consist of double skip connections~\cite{liu2020bi}, PReLU activations~\cite{bulat2019improved} and operation reorder (namely inserting a BatchNorm layer before binary activation)~\cite{rastegari2016xnor}. Different from previous works keeping $1 \times 1$ downsampling layers in full-precision for preserving the performance, we replace them all with max pooling operation, thus further reducing floating-point parameters and computing cost.

\noindent\textbf{Training Optimization.} In addition to network architectures, the training optimization of binary networks also requires a specialized and elaborate design. We use different optimization strategies for different datasets, and the detailed settings will be described in the following section.

\noindent\textbf{Activation Binarization.} Great efforts have been made on designing better binary activation by pioneering works~\cite{darabi2018bnn+,liu2020bi,wang2020sparsity}. In this work, we use the most primitive binary activation method as Xnor-Net~\cite{bulat2019xnor-net-plus}. Concretely, we use $sign(\cdot)$ for forward propagation and STE for backward propagation where $hard\_tanh(\cdot)$ is selected for gradient approximation.
\subsection{Ablation Study}
In the section above, the results of simply minimizing feature reconstruction error in Table~\ref{reconstruction_error} indicates that how to utilize the abundant features generated by $W$ needs careful design. In this section, we further present an ablation analysis on each object function in our label-aware representation approximation and report quantitative results in Table~\ref{tab:accuracy_of_three}. The baseline are obtained by compact ResNet-18 with Bi-Real~\cite{liu2020bi} and PReLU~\cite{bulat2019xnor-net-plus}. The quantitative results are consistent with our visualization Fig.~\ref{fig:visual_representation}. The instance-level approximation aligns the representations to present a similar distribution pattern. Based on it, the category-level approximation pulls the aligned representations together in pairs, making representations with the same label cluster more tightly. Since we add the $W$ to the computation graph, our performance is improved at the cost of approximately $1.33\times$ training time than the baseline. However, no extra computations are needed when deploying the trained binary models to the resource-limited devices.

\begin{table}[H]
\caption{Top-1 accuracies of compact ResNet-18 
 trained with three strategies. Both results evaluated with real-valued latent weights and binary weights are reported.}
\label{tab:accuracy_of_three}
\begin{center}
\begin{tabular}{lcc}
\toprule
\multirow{2}{*}{Training Strategy} &\multicolumn{2}{c}{Acc.($\%$)}\\
\cmidrule(lr){2-3}
&W &B\\
\toprule
Baseline & 41.13 & 82.37\\
\hline
Baseline + Eq.~\ref{eq:representation_approximation} & 81.60 & 82.26\\
\hline
Baseline + Eq.~\ref{eq:new_representation_approximation2} &84.77  & 85.10\\
\toprule
\end{tabular}
\vspace{-3mm}
\end{center}
\end{table}
We then evaluate the effect of the penalty coefficient $\lambda$.
A large penalty may reduce the performance of the network because it causes the total training objective to focus too much on the regular term at the expense of the original cross-entropy.
As shown in Table~\ref{tab:ablation_study_lamda}, a proper $\lambda$ matters in the balance between cross-entropy loss and the auxiliary term.
Since our specific supervision relies on the output features $\widetilde{Y}, Y$ of the backbone network, optimal $\lambda$ may change with different model architectures. Fortunately, $1e-4$ seems to perform well in most cases.
\begin{table}[h]
\caption{
Error rates ($\%$) on CIFAR-10 with compact ResNet-18 and WRN22, using different $\lambda$.}
\label{tab:ablation_study_lamda}
\begin{center}
\begin{tabular}{ccccc}
\toprule
\multirow{2}{*}{Model} &\multicolumn{4}{c}{$\lambda$}\\
\cmidrule(lr){2-5}
&1e-5 &1e-4 &1e-3 &1e-2 \\
\toprule
ResNet-18 &15.11 &14.90 &15.73  &17.09 \\
WRN-22 &6.89  &6.70 &7.35 &8.00\\
\toprule
\end{tabular}
\vspace{-3mm}
\end{center}
\end{table}

We find that a single linear fully-connected layer which projects penultimate representations to a learned space benefits our supervision.
Concretely, we project $\widetilde{Y}$ and $Y$ to a learned space with fixed dimension $D$ and then normalize them by dividing their own $\ell_2$ norm.
We empirically set the dimension $D$ of the new vector as 128 for ResNet-18/34 on ImageNet and 32 for small models on CIFAR datasets.
Empirical results with compact ResNet-18 on CIFAR-10 are reported in Table~\ref{tab:ablation_study_proj}. The improvement may be because that $\widetilde{Y}$ and $Y$ from binary space and latent space can be uniformed to a learned space with the same projection function. Based on Table~\ref{tab:ablation_study_lamda} and~\ref{tab:ablation_study_proj} We apply the best settings to the following experiments.
\begin{table}[h]
\begin{center}
\caption{Effect of using projection layer. `w/o' indicates without a projection layer.}
\label{tab:ablation_study_proj}
\begin{tabular}{c|cc}
\toprule
Projection Layer &with             &w/o  \\
\toprule
Error (\%)      &14.90($\pm$0.12) &15.95($\pm$0.25)\\
\toprule
\end{tabular}
\vspace{-5mm}
\end{center}
\end{table}
\subsection{Results}
In this section, we explore the effect of the proposed method from a quantitative view, by comparing it with the state-of-the-art low-bit networks on various architectures.

\noindent\textbf{CIFAR-10}
In all CIFAR experiments, we pad 2 pixels in each side of images and randomly crop $32\times32$ size from padded images during training. We use a batch size of 128 for training. Adam is adopted as the optimizer. The initial learning rate is set to $0.005$. The weight decay is $1e-6$. All networks are trained for $300$ epochs with cosine learning rate decay scheduler.

Table~\ref{tab:cifar10} shows our results compared with the state-of-the-art on CIFAR10 with various architectures including VGG, ResNet and WRN. The result is consistent with that on ImageNet, and our approach outperforms previous binarization methods like BNN and XNOR by a large margin.
\begin{table}[h]
    \caption{Performance comparison with SOTA methods on CIFAR-10.}
    \label{tab:cifar10}
	\centering
    \begin{tabular}{llcc}
		\toprule
		Models&Method&W/A&Acc.(\%)\\
		\midrule
		\multirow{6}{*}{VGG-small}&FP&32/32&91.7\\
        \cline{2-4}
		&BNN~\cite{NIPS2016_d8330f85}&1/1&89.9\\
		&XNOR~\cite{rastegari2016xnor}&1/1&89.8\\
		&Si-BNN~\cite{wang2020sparsity}&1/1&90.2\\
		&IR-Net~\cite{qin2020forward}&1/1&90.4\\
		&\textbf{Ours}&{1/1}&\textbf{91.2}\\
        \midrule
        \multirow{4}{*}{\shortstack[l]{ResNet-18\\(16-16-32-64)}}&FP&32/32&90.77\\
        \cline{2-4}
		&PCNN~\cite{gu2019projection}&1/1&78.93\\
		&CBCN~\cite{liu2019circulant}&1/1&80.32\\
		&\textbf{Ours}&1/1&\textbf{85.10}\\
		\midrule
		\multirow{4}{*}{ResNet-20}&FP&32/32&91.7\\
        \cline{2-4}
		&IR-Net~\cite{qin2020forward}\cite{qin2020forward}&1/1&85.4\\
		&RBNN~\cite{lin2020rotated}&1/1&87.8\\
		&\textbf{Ours}&{1/1}&\textbf{88.6}\\
		\midrule
        \multirow{4}{*}{WRN-22}&FP&32/32&95.75\\
        \cline{2-4}
		&PCNN~\cite{gu2019projection}&1/1&91.37\\
        &BONN~\cite{gu2019bayesian}&1/1&92.36\\
		&\textbf{Ours}&1/1&\textbf{93.30}\\
		\bottomrule
	\end{tabular}
\end{table}

\noindent\textbf{ImageNet} For the large-scale dataset, we evaluate our approach over ResNet-18/34 on ImageNet. As for data preprocessing, we first proportionally resize images to $256\times N (N \times 256)$ with the short edge to $256$. Then we randomly sub-crop them to $224\times 224$ patches with mean subtraction and randomly flipping. No other data augmentation tricks are used during training. Similar to settings on CIFAR, we use Adam with cosine learning rate. The initial learning rate is set to $0.001$ and the weight decay is set to $0$. All networks are trained from scratch for $120$ epochs.

We compare our method with several exiting state-of-the-art extreme low-bit quantization methods that binarize both weights and activations: Bi-Real~\cite{liu2020bi}, PCNN~\cite{gu2019projection}, Si-BNN~\cite{wang2020sparsity}, CI-BNN~\cite{Wang_2019_CVPR}, IR-Net~\cite{qin2020forward}, RBNN~\cite{lin2020rotated} and SA-BNN~\cite{LiuBNN2021AAAI}. The overall results are shown in Tabel~\ref{resnet}.
In detail, using ResNet-18 with Bi-Real~\cite{liu2020bi}+PReLU~\cite{bulat2019improved} architecture, our method achieves 63.4$\%$ in Top-1 accuracy. Besides, we highlight our Top-1 accuracy on ResNet-34 is $67.0\%$, which is an up to 1.5$\%$ absolute improvement compared with state-of-the-art SA-BNN (both with Bi-Real+PReLU architecture). 
The results show that our training method outperforms the best previous binary methods. In addition, unlike previous works keeping $1\times 1$ downsampling layers in full-precision, we replace them all with max pooling. Therefore, our models can benefit from fewer Flops with the same Bops when deployed to source-limited devices.

\subsection{Training and inference complexity analysis}
During training, the forward propagation time and backward propagation time is in the proportion $1:2$.
Since we add the $W$ to the computation graph only for forward propagation, our performance is improved at the cost of approximately $1.33\times$ training time than our baseline. Even so, the total training time is still less than previous two-stage training works~\cite{liu2020bi,martinez2020training,han2020training} that need pre-training real-valued models at first.

As for the inference, no extra computations are needed when deploying the trained binary models to the resource-limited devices. Besides, we replace all full-precision $1 \times 1$ downsampling layers with max pooling operation. Therefore, our models can benefit from fewer Flops with the same Bops when deployed to source-limited devices. The detailed inference complexity analysis on ResNet-18/34 including memory usage, Flops and Bops is reported in Table~\ref{resnet}.
\begin{table*}[th]
\caption{Memory usage, Flops and Bops analysis at inference time. Unlike previous works keeping $1 \times 1$ downsampling layers in full-precision, we replace them all with max pooling to further reduce real-valued Flops.}
\label{resnet}
\begin{center}
\begin{tabular}
{llcccc}
\toprule
Models & Methods  &Memory & Ratio & Flops &Bops  \\
\toprule
\multirow{5}{*}{ResNet-18}
& FP                                &374.1 Mbit & 1$\times$ & $1.826\times10^9$ &0 \\
& Xnor-Net \cite{rastegari2016xnor} &33.7 Mbit & 11.10$\times$ & 1.498 $\times10^8$ &$1.676 \times 10^9$\\
& SA-BNN \cite{LiuBNN2021AAAI}      &33.6 Mbit & 11.14$\times$ & 1.513 $\times10^8$ &$1.676 \times 10^9$\\
& Bi-Real \cite{liu2020bi}          &33.6 Mbit & 11.14$\times$ & 1.393 $\times10^8$ &$1.676 \times 10^9$\\
& Ours                              &33.1 Mbit &11.30$\times$ &\textbf{$1.210\times 10^8$} &$1.676 \times 10^9$ \\
\hline\hline
\multirow{5}{*}{ResNet-34}
& FP                                &697.3 Mbit & 1$\times$ & $3.667\times10^9$ &0 \\
& Xnor-Net \cite{rastegari2016xnor} &43.9 Mbit & 15.88$\times$ & 1.618 $\times10^8$ &$3.526 \times 10^9$\\
& SA-BNN \cite{LiuBNN2021AAAI}      &44.1 Mbit & 15.81$\times$ & 1.648 $\times10^8$ &$3.526 \times 10^9$\\
& Bi-Real \cite{liu2020bi}          &43.7 Mbit & 15.97$\times$ & 1.408 $\times10^8$ &$3.526 \times 10^9$\\
& Ours                              &43.1 Mbit &16.17$\times$ &\textbf{$1.247\times 10^8$} &$3.526 \times 10^9$ \\
\toprule
\end{tabular}
\end{center}
\end{table*}

\subsection{Discussion}
\begin{table}[t]
\caption{Comparison of accuracy ($\%$) on ImageNet with state-of-the-art methods in ResNet-18/34. `KD' means that we combine our method with knowledge distillation.}
\label{resnet}
\begin{center}
\begin{tabular}
{llccc}
\toprule
Models & Methods  &W/A & Top-1 & Top-5  \\
\toprule
\multirow{10}{*}{ResNet-18} & Bi-Real \cite{liu2020bi} &1/1 & 56.4 & 79.5 \\
& PCNN \cite{gu2019projection} &1/1 & 57.3 & 80.0  \\
& Si-BNN \cite{wang2020sparsity} &1/1 & 59.7 & 81.8  \\
& CI-BNN \cite{Wang_2019_CVPR} &1/1 & 59.9 & 84.2  \\
& IR-Net \cite{qin2020forward} &1/1 & 58.1 & 80.0  \\
& RBNN \cite{lin2020rotated} &1/1 & 59.9 & 81.9  \\
& SA-BNN \cite{LiuBNN2021AAAI} &1/1 & 61.7 & 82.8  \\
& \textbf{Ours}  &1/1& \textbf{63.4} &\textbf{84.6}  \\
& \textbf{Ours+KD}  &1/1& \textbf{63.8} & \textbf{84.9}  \\
\cline{2-5}
& FP32 &32/32 & 69.3 & 89.2\\
\hline\hline
\multirow{9}{*}{ResNet-34} 
& Bi-Real \cite{liu2020bi} &1/1 & 62.2 & 83.9 \\
& IR-Net \cite{qin2020forward} &1/1 & 62.9 & 84.1 \\
& Si-BNN \cite{wang2020sparsity}  &1/1 & 63.3 & 84.4 \\
& CI-BNN \cite{Wang_2019_CVPR} &1/1 & 64.9 & 86.6  \\
& RBNN \cite{lin2020rotated} &1/1 & 63.1 & 84.4  \\
& SA-BNN \cite{LiuBNN2021AAAI} &1/1 & 65.5 & 85.8  \\
& \textbf{Ours}  &1/1 & \textbf{67.0} & \textbf{86.8} \\
& \textbf{Ours+KD}  &1/1 & \textbf{67.3} & \textbf{87.0} \\
\cline{2-5}
& FP32 &32/32 & 73.3 & 91.3 \\
\toprule
\end{tabular}
\vspace{-2mm}
\end{center}
\end{table}
In our approach, we utilize float-precision $W$ to assist the training of binary networks, which sounds similar to the popular knowledge distillation (KD). However, there are actually significant differences: (1) The latent weight $W$, as shown in Fig.~\ref{fig:pipeline}, has always existed as an auxiliary variable in the original binary training framework rather than being introduced additionally. It is different from KD that introduces additional well-trained large teacher models. Thus there is no overhead in choosing proper teacher models.
(2) Unlike KD using a well-trained large teacher and keeping constant at training, the auxiliary latent variable $W$ and binary variable $B$ are optimized simultaneously in our approach. They are related as Eq.~\ref{WandB} and are always updated synchronously throughout training, providing different representations but at the same stage of convergence as Eq.~\ref{eq:inference_w}.
(3) In addition, the role of the teacher model in KD is to guide the training by providing soft targets, while $W$ in our method serves for label-aware representation approximation, which provides distinctive information with KD. Therefore, our approach can be combined with KD to improve performance further.

Based on the above discussion, we further combine our approach with KD-based method Label Refinery~\cite{bagherinezhad2018label} (LR). Concretely, we choose well-trained ResNet-34 from model zoo as a teacher model and simply replace the CE loss $\mathcal{L}_{ce}=\sum_{k=1}^K -y_k\cdot \log (p_k)$ in Eq.~\ref{eq:loss} with Label Refinery loss $\mathcal{L}_{LR}=\sum_{k=1}^K -y_k\cdot \log (T_k)$ where $p_k$ is the one-hot label distribution and $T_k$ is the soft logit distribution from teacher model. The results on ImageNet are shown in Table.~\ref{resnet} and the combined training further improves the Top-1 accuracy of ResNet18 and ResNet34 by $0.4\%$ and 0.3$\%$, respectively.
\section{Conclusion}
In this work, we explore the role of the latent weight besides weight approximation and propose to add it to the computation graph. With appropriate processing, we restore the latent weight's feature extraction ability to aid the binary training. We show that our label-aware representation approximation not only allows representations of $W$/$B$ to present a similar distribution pattern, but also allows them with the same label cluster more tightly. The effectiveness is verified through qualitative and quantitative experiments.

{\small
\bibliographystyle{ieee_fullname}
\bibliography{egbib}
}

\end{document}